\pdfoutput=1

\documentclass[10pt,twocolumn,letterpaper]{article}

\usepackage[pagenumbers]{wacv}  

\usepackage{algorithm}
\usepackage{algpseudocode}
\usepackage{tikz}
\usetikzlibrary{arrows.meta, positioning, calc}
\usepackage{multirow}

\newcommand{\ncgrpo}{NC-GRPO}

\newtheorem{proposition}{Proposition}
\newtheorem{lemma}{Lemma}

\newenvironment{proofof}[1]{\noindent\emph{Proof of #1.}\hspace{0.5em}}{\hfill$\square$\par\medskip}

\newcommand{\E}{\mathbb{E}}
\newcommand{\R}{\mathbb{R}}
\newcommand{\Var}{\mathrm{Var}}
\newcommand{\tr}{\mathrm{tr}\,}
\newcommand{\norm}[1]{\lVert #1 \rVert}

\newcommand{\TODO}[1]{\textbf{\color{red}[TODO: #1]}}

\renewcommand{\TODO}[1]{}

\definecolor{wacvblue}{rgb}{0.21,0.49,0.74}
\usepackage[pagebackref,breaklinks,colorlinks,allcolors=wacvblue]{hyperref}

\def\confName{WACV}
\def\confYear{2027}

\title{Perturb the Thought, Not the Pixels: Latent-Space Rollout Diversification for Reinforcement Learning of Vision--Language Models}

\author{Michael Jerge\\
Amazon Web Services\\
{\tt\small mjerge@amazon.com}
\and
Joseph Pelczar\\
Amazon Web Services\\
{\tt\small jpelczar@amazon.com}
\and
Justin Downes\\
Amazon Web Services\\
{\tt\small jusdow@amazon.com}
}

\begin{document}
\maketitle
\begin{abstract}
Reinforcement learning with verifiable rewards (RLVR) improves the reasoning ability of vision--language models (VLMs), and diversifying the rollouts within each optimization group amplifies its gains. Existing approaches diversify through decoding temperature or pixel-space image distortion; we ask whether the perturbation belongs in the model's latent space instead. We introduce Noise-Contrastive GRPO (NC-GRPO), which injects scale-calibrated Gaussian noise into the last hidden layer of the prompt-encoding pass for half of each rollout group, branching those rollouts from a displaced departure state. Branches that reach the answer despite the displacement are reinforced over those derailed by it, converting sensitivity at the branch point into policy-gradient signal; the objective, reward, and inference protocol are untouched. On Qwen2.5-VL-7B trained on Geometry3K, NC-GRPO significantly improves out-of-domain mathematical reasoning over vanilla GRPO across five held-out benchmarks (pooled McNemar $p \le 0.001$) while also improving in-domain accuracy and hallucination robustness --- the latter an axis on which image-space noise regresses even while posting a larger OOD average on perception-heavy benchmarks. Mechanism ablations indicate that independent stochastic diversity, not noise budget or direction, is the active ingredient, and a noise-scale study exposes a dial between reasoning specialization and general capability. NC-GRPO is designed to be modality-agnostic and integrates into a standard RLVR pipeline as a ${\sim}50$-line change to the inference engine.
\end{abstract}


\section{Introduction}
\label{sec:intro}

Reinforcement learning from rewards has become a central paradigm for improving the reasoning capabilities of large language models \cite{deepseekr1,ouyang2022instructgpt}. Group Relative Policy Optimization (GRPO) \cite{shao2024deepseekmath} generates a group of rollouts per prompt, normalizes rewards within the group, and trains the policy to favor rollouts with above-average reward. A key design choice is how to generate diverse rollouts: if all rollouts in a group agree, the group carries no learning signal. The standard mechanism is temperature sampling from the current policy \cite{vonklinski2026taxonrl}; NoisyRollout \cite{liu2025noisyrollout} showed that additionally conditioning half of each group on a Gaussian-distorted copy of the input image --- with the distortion annealed over training --- improves out-of-domain (OOD) generalization on multimodal reasoning benchmarks.

Both mechanisms share a limitation: a model whose answer is encoded ambiguously in its hidden states may, in principle, produce token-diverse samples under temperature that cluster around the same incorrect latent mode, and image-space distortion can only diversify along directions the visual encoder can express --- and cannot be applied to text-only models at all. Neither acts directly on the model's internal problem representation, motivating a simple question: does the perturbation belong in the model's latent space instead?

\paragraph{Our approach.} Rather than diversifying rollouts through decoding stochasticity or pixel distortion, we inject Gaussian noise directly into the hidden representations during the prompt-encoding (prefill) forward of the noisy half of each group (workflow diagram in the supplementary material). Each noisy branch thus departs from a displaced point in representation space; branches that reach the answer despite the displacement are reinforced relative to branches derailed by it, and the group-normalized advantage converts insensitivity at the branch point into a policy-gradient signal. We are precise in \cref{sec:branching} about scope: the noise displaces the state each rollout departs from, without altering the cached representation it attends to while generating. Everything else --- objective, reward, decoding, evaluation --- is identical to GRPO, yielding a clean three-way comparison between vanilla GRPO, image-space diversification, and latent-space diversification.

Latent perturbation has three a-priori advantages: it is \emph{modality-agnostic by design} (defined for any model with hidden representations, including text-only LLMs, though validated here only on VLMs); \emph{cheap by construction} (no second image-processing pass, no re-encoding of the vision tower); and \emph{controllable} --- as a vector in representation space it can be negated (antithetic pairs), steered (learned directions), and scaled exactly (\cref{sec:probes}), letting us dissect not just \emph{whether} latent noise helps but \emph{what} property of the noise is responsible.

\paragraph{Summary of contributions.}
\begin{itemize}
  \item \textbf{Method:} NC-GRPO, which replaces pixel-space rollout diversification with noise-induced branching in representation space, implemented as a seed-synchronized hidden-state hook in the inference engine (\cref{sec:method}).
  \item \textbf{Results:} significant OOD reasoning gains over vanilla GRPO on five multimodal math benchmarks (pooled McNemar $p{=}0.001$ -- $p{<}10^{-4}$ across noise scales), improved in-domain accuracy, and better hallucination robustness than image-space noise (significant on POPE, $p{=}0.007$; HallusionBench $+0.6$pp) (\cref{sec:results,sec:imagecomparison}).
  \item \textbf{Mechanism:} matched antithetic and steered probes showing that independent stochastic diversity --- not noise budget or direction --- is the active ingredient (\cref{sec:mechanism}).
  \item \textbf{Specialization dial:} a noise-scale study showing that $\sigma_0$ trades general capability for reasoning gains, with a regime ($\sigma_0{=}0.2$) that improves reasoning at no measurable general-capability cost (\cref{sec:imagecomparison}).
\end{itemize}

\section{Background and Related Work}
\label{sec:background}

\subsection{Group Relative Policy Optimization}
\label{sec:grpo}

GRPO \cite{shao2024deepseekmath} optimizes a policy $\pi_\theta$ by generating a group of $G$ rollouts $\{o_i\}_{i=1}^{G}$ for each prompt $q$, computing verifiable rewards $\{r_i\}_{i=1}^{G}$, and applying the clipped surrogate objective
\begin{multline}
\mathcal{L}_{\mathrm{GRPO}} = -\E\Bigg[ \min\!\Bigg( \frac{\pi_\theta(o \mid q)}{\pi_{\theta_{\mathrm{old}}}(o \mid q)}\,\hat{A},\\
\operatorname{clip}\!\Big( \frac{\pi_\theta(o \mid q)}{\pi_{\theta_{\mathrm{old}}}(o \mid q)}, 1 \pm \varepsilon \Big) \hat{A} \Bigg) + \beta\, \mathrm{KL}\big[\pi_\theta \,\|\, \pi_{\mathrm{ref}}\big] \Bigg],
\label{eq:grpo}
\end{multline}
where the advantage is normalized within the group, $\hat{A}_i = (r_i - \bar{r})/(\operatorname{std}(r) + \delta)$. GRPO descends from PPO \cite{schulman2017ppo}, with successors refining the objective \cite{deepseekr1,yu2025dapo,liu2025drgrpo}; RLHF was introduced by \cite{christiano2017deep} and scaled to instruction-following LLMs with PPO by \cite{ouyang2022instructgpt}. Applied to VLMs, RLVR improves multimodal mathematical reasoning \cite{huang2025visionr1,meng2025mmeureka,deng2025openvlthinker}; prior GRPO-VLM work generates rollout diversity through temperature sampling \cite{vonklinski2026taxonrl} --- the output-space mechanism NC-GRPO replaces --- and the on-policy rollout distribution has been shown to determine what preference optimization can teach a VLM about hallucination \cite{yu2025onpolicy}. Our work modifies only rollout generation and is complementary to objective and reward-model improvements.

\subsection{Rollout diversification}
\label{sec:noisyrollout}

NoisyRollout \cite{liu2025noisyrollout} is the direct predecessor of this work and supplies the scaffolding we build on: it conditions the noisy half of each group on a Gaussian-distorted copy of the input image, anneals the distortion to zero with a sigmoid schedule, and leaves the objective unchanged. Each group contains $n$ clean-conditioned and $n$ perturbed-conditioned rollouts ($G = 2n$), with advantages computed over the joint group, so systematic disagreement between the halves is itself learning signal. We inherit the two-pass structure, 50/50 split, schedule, and training protocol exactly, changing only the perturbation site --- from pixels to hidden states --- which makes the image-vs-latent comparison a controlled experiment on a pipeline the community already uses. The two perturbations are not nested: a pixel distortion propagates into the representation every generated token attends to, whereas ours displaces only the branch point, isotropically and controllably (\cref{sec:branching}). Our injection site follows NoisyCoconut \cite{jerge2026noisycoconut}, which perturbs the last hidden layer of a first forward pass to branch reasoning paths from a shared state for inference-time confidence estimation; NC-GRPO moves that construction into the RLVR training loop, where the divergence is consumed by the group-relative advantage rather than an aggregation rule.

\subsection{Uncertainty and latent perturbation}
\label{sec:latentrelated}

Monte Carlo dropout \cite{gal2016dropout} and deep ensembles \cite{lakshminarayanan2017ensembles} estimate uncertainty through stochastic forward passes and model diversity; our branches provide representation-level diversity within one model, used for training rather than inference. Hidden-state perturbation also appears as a regularizer (R-Drop \cite{liang2021rdrop}), as exploration in parameter/latent space (evolution strategies \cite{salimans2017es}, SPSA \cite{spall1992spsa}), in flat-minima optimization (SAM \cite{foret2021sam}), and as a medium for reasoning (Coconut \cite{hao2024coconut}). Closest in spirit, NEFTune \cite{jain2024neftune} adds uniform noise to token embeddings during supervised fine-tuning --- but NC-GRPO is not NEFTune inside GRPO: NEFTune perturbs every training example under a likelihood loss, where the noise regularizes the objective itself, whereas here noise enters only rollout \emph{generation} (loss, log-probabilities, and gradients see unperturbed states) and only half of each group, so the effect is routed entirely through the group-relative advantage contrast (\cref{prop:decomposition}) --- a mechanism with no analogue in supervised fine-tuning. Our contribution is placing latent perturbation inside the RLVR rollout loop and causally dissecting its mechanism.

\section{Method: NC-GRPO}
\label{sec:method}

\subsection{Noise-induced branching}
\label{sec:branching}

Given hidden states $h_\ell$ at transformer layer $\ell$, the general form of our perturbation is
\begin{equation}
h_\ell^{(i)} = h_\ell + \epsilon_\ell^{(i)}, \qquad \epsilon_\ell^{(i)} \sim \mathcal{N}\big(0, \sigma_\ell^2 I\big),
\label{eq:general}
\end{equation}
applied to the hidden states returned by the prefill (first) forward pass of branch $i$; each branch then decodes without further noise, so decode cost is identical to standard GRPO.

We instantiate \cref{eq:general} at the last hidden layer, following NoisyCoconut \cite{jerge2026noisycoconut}, which injects Gaussian noise into the last hidden layer of a language model's first forward pass in order to branch multiple reasoning paths from a common initial state. NoisyCoconut's formulation iterates $h_{t+1} = f_\theta(h_t + \eta_t)$, feeding the perturbed state forward across latent reasoning steps; a standard autoregressive decode has no recurrent latent state to iterate, and the single-shot prefill injection is its faithful adaptation to that setting --- the noise seeds divergent paths from a common perturbed initial state. NC-GRPO places that construction inside the RLVR rollout loop and reads the resulting divergence through the group-relative advantage. The scale is calibrated to the hidden norm so that $\sigma$ is dimensionless and width-invariant. For prompt token $t$ with hidden state $h_t \in \R^d$:
\begin{equation}
\tilde{h}_t = h_t + \sigma_k \cdot \frac{\norm{h_t}}{\sqrt{d}} \cdot \epsilon_t, \qquad \epsilon_t \sim \mathcal{N}(0, I_d),
\label{eq:calibrated}
\end{equation}
where $\sigma_k$ follows the annealing schedule of NoisyRollout \cite{liu2025noisyrollout},
\begin{equation}
\sigma_k = \sigma_0 \left( 1 - \operatorname{sigmoid}\!\Big( \gamma\, \tfrac{k - k_{\mathrm{mid}}}{K} \Big) \right),
\label{eq:schedule}
\end{equation}
over training steps $k \in \{1, \ldots, K\}$: strong exploration early, annealed to zero as the policy converges. The norm calibration in \cref{eq:calibrated} makes $\sigma_k$ an exact, width-invariant relative perturbation:

\begin{proposition}[Scale calibration]
\label{prop:calibration}
Under \cref{eq:calibrated}, $\E \| \tilde{h}_t - h_t \|^2 = \sigma_k^2 \norm{h_t}^2$ for every hidden width $d$, and the relative perturbation concentrates: $\Var\big( \| \tilde{h}_t - h_t \|^2 / (\sigma_k^2 \norm{h_t}^2) \big) = 2/d$. (Proof in the supplementary material.)
\end{proposition}

When $d$ is in the thousands, the relative perturbation magnitude thus concentrates tightly around $\sigma_k \norm{h_t}$ (relative variance $2/d$). Note that \cref{prop:calibration} fixes the Euclidean size of the perturbation, not its behavioral effect; whether a fixed $\sigma_0$ induces comparable behavioral change across models is an empirical question we address, within one model line, in \cref{sec:generality}. Noise is applied only during rollout generation, and evaluation is always noise-free; under tensor parallelism a step-seeded generator makes all ranks perturb replicated hidden states identically. Log-probabilities are computed without noise, so the importance ratio in \cref{eq:grpo} is evaluated exactly; as in NoisyRollout \cite{liu2025noisyrollout}, the perturbed half constitutes an off-policy proposal for which we apply no density correction. More general instantiations of \cref{eq:general} --- multi-layer injection with depth-decayed scales and a learned noise head --- are natural extensions we do not evaluate here; all headline results use \cref{eq:calibrated}.

\paragraph{Where the perturbation acts.} The hook fires on the tensor \emph{returned} by the language-model stack, after every layer has run; the KV cache is written inside that call from unperturbed activations, and the returned tensor's only consumer is the logit computation. The perturbation therefore shifts the logit distribution at the final prompt position --- the first generated token --- after which each branch decodes from a clean cache (during decode the hook is inert). \Cref{eq:calibrated} is thus a \emph{branch-point} intervention by design, not an engine-imposed approximation: it displaces the state a rollout departs from, and autoregression carries the divergence forward, exactly NoisyCoconut's branching construction. It does not perturb the representation the rollout attends to while generating --- reaching the KV cache would require injecting at or before a layer's input, and that propagating variant, sustained rather than seeded, is a companion experiment whose mechanism we do not claim (\cref{sec:limitations}). The mechanism is therefore closer to structured branching at the point of divergence than to sustained representational noise: distinct from raising decode temperature, since the displacement passes through the LM head as a correlated, per-branch-seeded shift in logit space rather than a uniform rescaling of it, but nearer to first-token diversification than a latent-manifold description would suggest.

\subsection{What the mixed group learns: an advantage decomposition}
\label{sec:decomposition}

Why should placing clean and noisy rollouts in one advantage group create a robustness signal, rather than merely adding sampling noise? The group-normalized advantage makes this explicit.

\begin{proposition}[Mixed-group advantage decomposition]
\label{prop:decomposition}
Let a group contain $n$ clean rollouts with rewards $\{r_i^{c}\}$ (mean $\bar{r}^{c}$) and $n$ noisy rollouts with rewards $\{r_j^{\eta}\}$ (mean $\bar{r}^{\eta}$), and let $s$ denote the group reward standard deviation. Then the advantage of every clean rollout satisfies
\begin{equation}
\hat{A}_i^{c} = \underbrace{\frac{r_i^{c} - \bar{r}^{c}}{s}}_{\text{within-half}} + \underbrace{\frac{\bar{r}^{c} - \bar{r}^{\eta}}{2s}}_{\text{contrast}}, \quad
\hat{A}_j^{\eta} = \frac{r_j^{\eta} - \bar{r}^{\eta}}{s} - \frac{\bar{r}^{c} - \bar{r}^{\eta}}{2s},
\label{eq:advantage}
\end{equation}
i.e., every advantage is the ordinary within-half advantage plus a contrast term of equal magnitude and opposite sign across the two halves, which vanishes iff the policy's expected reward is invariant to the perturbation. (Proof in the supplementary material.)
\end{proposition}

The contrast term is what ordinary GRPO lacks: whenever perturbing the encoding changes expected reward, the gradient uniformly reinforces the surviving half against the derailed one --- a pressure toward policies whose reward is insensitive to representational perturbation. \Cref{sec:agreement} makes precise what that insensitivity buys; the antithetic probe (\cref{sec:probes}) tests cancelling the stochastic component of this very term.

\subsection{Why should this help? An agreement view}
\label{sec:agreement}

The mechanism can be understood through what we term \emph{representational agreement} --- the consistency of a single model's internal state under perturbation (distinct from CKA/RSA \cite{kornblith2019cka}, which compare across models or layers). Define the agreement between two perturbed branches via an L2-Gaussian kernel,
\begin{equation}
\kappa\big(h^{(i)}, h^{(j)}\big) = \exp\!\left( - \frac{\big\| h^{(i)} - h^{(j)} \big\|^2}{2\tau^2} \right),
\label{eq:agreement}
\end{equation}
with bandwidth $\tau$ the RMS of pairwise distances. A question whose final prefill state is stable yields noisy rollouts that match the clean ones, and the group advantage is driven by ordinary reward variation; a fragile state yields diverging noisy rollouts --- and because advantages are normalized over the joint group (\cref{eq:grpo}), correct-and-stable behavior is reinforced precisely where the departure state is fragile. Aggregated over training, this selects for policies whose reward is insensitive to displacement of that state, a restricted analogue of flat-minima regularization \cite{foret2021sam} whose scope is set by the injection site (\cref{sec:branching}): the contrast term selects for insensitivity to perturbation of the final prefill state, whose influence runs entirely through the first generated token --- not for robustness to perturbation sustained across generation, which our intervention never applies.

Concretely, for smooth expected reward $\varphi(h) = \E[\, r \mid h \,]$, a second-order expansion under \cref{eq:calibrated} gives
\begin{equation}
\E_\epsilon\big[ \varphi(\tilde{h}) \big] = \varphi(h) + \frac{\sigma^2 \norm{h}^2}{2}\, \bar{\lambda}\big( \nabla^2 \varphi(h) \big) + O(\sigma^3),
\label{eq:smoothing}
\end{equation}
with $\bar{\lambda}$ the mean Hessian eigenvalue (proof in the supplementary material). Near a maximum $\bar{\lambda} \le 0$: the contrast term of \cref{prop:decomposition} is a per-prompt curvature estimate at the branch point, and the update ascends the Gaussian-smoothed reward, preferring flat maxima --- tested causally in \cref{sec:mechanism}.

\subsection{The geometry of input-space vs.\ latent-space perturbation}
\label{sec:geometry}

The two noise families explore geometrically different representations. To first order, pixel noise induces a perturbation with covariance $s^2 J J^\top$ ($J$ the encoder Jacobian): supported on $\operatorname{range}(J)$, input-dependent in power, with a smoothing penalty filtered through the encoder geometry (formal statement in the supplementary material); latent noise perturbs all of $\R^d$ isotropically with exact calibration (\cref{prop:calibration}). This predicts the division of labor in \cref{sec:imagecomparison}: pixel noise wins on perception (WeMath, MathVerse), latent noise on hallucination and in-domain accuracy. Two asymmetries run the other way: pixel noise explores only encoder-consistent perturbations (the right bias for perception robustness), and it is the more \emph{persistent} intervention, reaching the keys and values every token attends to while \cref{eq:calibrated} displaces only the branch point (\cref{sec:branching}) --- so the contrast is isotropic branch-point displacement vs.\ encoder-filtered sustained perturbation, not deep vs.\ shallow, and not in latent noise's favor on persistence. What latent space uniquely affords is algebraic structure --- exact negation, scaling, and direction, on which every probe in \cref{sec:probes} depends; pixel space supports none of these exactly (the ``negation'' of a distortion is undefined), so the controlled experiments of \cref{sec:mechanism} have no pixel-space counterpart.

\subsection{Mechanism probes: antithetic pairs and learned steering}
\label{sec:probes}

Because the perturbation is a vector in representation space, we can construct matched interventions that isolate \emph{what} about the noise matters.

\paragraph{Antithetic pairs.} Split the noisy half of the group into two quarter-groups sharing the same draw with opposite signs, $(+\epsilon, -\epsilon)$, at identical compute --- the classical antithetic-variates construction, whose effect on the mixed-group signal of \cref{prop:decomposition} is exact:

\begin{lemma}[Antithetic cancellation]
\label{lem:antithetic}
Write $\varphi = \varphi_{\mathrm{even}} + \varphi_{\mathrm{odd}}$ for the even/odd decomposition of the reward response $\varphi(\epsilon) = \E[\, r \mid \epsilon \,]$. Antithetic pairing leaves the marginal distribution of every rollout unchanged, but replaces the noisy-half mean by an average of pair-means $\frac{1}{2}(\varphi(\epsilon) + \varphi(-\epsilon)) = \varphi_{\mathrm{even}}(\epsilon)$: the odd component --- which contains the entire first-order response $\nabla \varphi(0)^\top \epsilon$, the dominant fluctuation at small $\sigma$ --- is removed exactly, while the $O(\sigma^2)$ even (curvature) component of \cref{eq:smoothing} is retained. (Proof in the supplementary material.)
\end{lemma}

If that first-order fluctuation were nuisance variance --- an estimation cost of the method --- pairing should help or be neutral; if the independent spread of noisy-rollout rewards is the training signal, cancelling it should hurt.

\paragraph{SPSA-learned steering.} With rank-one noise $\epsilon_t = r_t u$, where $u \in \R^d$ is a per-step direction shared across tokens and $r_t$ a scalar, the antithetic reward gap yields a two-point simultaneous-perturbation estimate of the reward gradient along $u$ \cite{spall1992spsa}:
\begin{equation}
\hat{g}_k = \frac{\bar{R}^{+} - \bar{R}^{-}}{2 \sigma_k}\, u_k, \qquad
\mu_{k+1} = \Pi_{\norm{\cdot} \le \mu_{\max}}\!\big( \mu_k + \eta\, \hat{g}_k \big),
\label{eq:spsa}
\end{equation}
accumulating a steering vector $\mu$ added to every noisy prefill ($\tilde{h}_t \mathrel{+}= \mu \cdot \norm{h_t} / \sqrt{d}$), converting undirected noise into a learned latent intervention \cite{salimans2017es}. If a beneficial fixed direction existed in representation space, steering should outperform undirected noise; if the benefit lives in the independent randomness itself, it should not.

The supplementary material lists the full training-step algorithm. The implementation is a ${\sim}50$-line, step-seeded hook on the hidden states returned by the inference engine's language-model stack, plus a two-pass generation call; no change to the trainer's loss, reward, or evaluation code. The same hook pattern is used for every model in \cref{sec:generality}, so the branch-point characterization above applies uniformly across them.

\section{Experiments}
\label{sec:experiments}

\subsection{Setup}
\label{sec:setup}

\paragraph{Training.} Qwen2.5-VL-7B-Instruct \cite{bai2025qwen25vl}, GRPO on Geometry3K \cite{lu2021intergps} (verifiable numeric/choice answers; rule-based reward on \verb|\boxed| content), 60 steps, batch 128 prompts, $n{=}6$ rollouts per pass ($G{=}12$ for two-pass methods), response budget 2048 tokens, frozen vision tower, vLLM rollouts. All methods share every hyperparameter; only the perturbation differs. Schedule (\cref{eq:schedule}): $\gamma{=}30$, $k_{\mathrm{mid}}{=}40$, matching \cite{liu2025noisyrollout}.

\paragraph{Evaluation.} Greedy decoding, identical rule-based scorer for every method, on in-domain Geometry3K (601 questions) and five OOD math benchmarks: HallusionBench \cite{guan2024hallusionbench}, MathVista \cite{lu2024mathvista}, MathVerse \cite{zhang2024mathverse}, MathVision \cite{wang2024mathvision}, WeMath \cite{qiao2024wemath} ($n{=}10{,}038$ OOD questions). We additionally evaluate general capability on POPE \cite{li2023pope} (object hallucination), MMStar \cite{chen2024mmstar}, AI2D \cite{kembhavi2016ai2d}, RealWorldQA \cite{xai2024realworldqa}, and ChartQA \cite{masry2022chartqa} (\cref{sec:imagecomparison}). All comparisons are per-question paired exact McNemar tests --- every method answers the same questions. Per-question records, training code, and the injection hook will be made publicly available.

\subsection{Main results}
\label{sec:results}

\begin{figure*}[t]
  \centering
  \includegraphics[width=\textwidth]{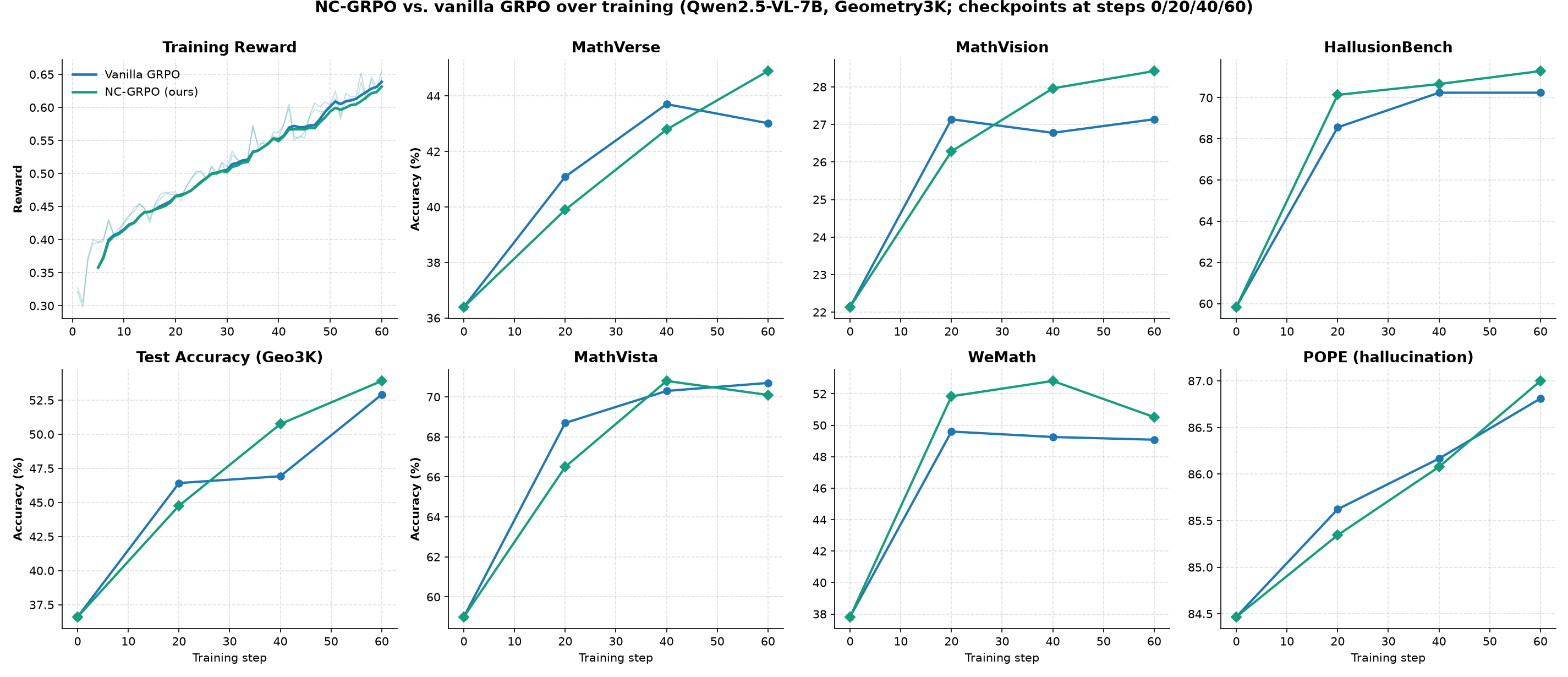}
  \caption{Training dynamics (Qwen2.5-VL-7B, Geometry3K, 60 steps; checkpoints at 0/20/40/60). Training reward (top left) matches vanilla step-for-step (mean difference $-0.003 \pm 0.005$; $\pm$ is std across steps, not seed spread) while held-out accuracy diverges: by step 60 NC-GRPO leads on four of five OOD benchmarks, in-domain accuracy, and POPE. Gains appear OOD, not in training reward --- convergence to a different, more robust solution (\cref{sec:agreement}).}
  \label{fig:dynamics}
\end{figure*}

\begin{table*}[t]
  \centering
  \small
  \begin{tabular}{lccccccc}
    \toprule
    Method & Geo3K (ID) & Hallusion & MathVista & MathVerse & MathVision & WeMath & OOD avg \\
    \midrule
    Vanilla GRPO           & 52.9 & 70.2 & 70.7 & 43.0 & 27.1 & 49.1 & 52.0 \\
    NC-GRPO $\sigma_0{=}0.2$ & \textbf{53.9} & \textbf{71.3} & 70.1 & 44.9$^*$ & \textbf{28.4} & 50.5 & 53.0$^*$ \\
    NC-GRPO $\sigma_0{=}0.5$ & 53.7 & 71.0 & \textbf{71.2} & \textbf{45.3}$^*$ & 27.6 & \textbf{52.1}$^*$ & \textbf{53.5}$^*$ \\
    \bottomrule
  \end{tabular}
  \caption{\textbf{Main result: reasoning benchmarks.} Accuracy (\%) vs.\ vanilla GRPO (Qwen2.5-VL-7B, Geometry3K, greedy eval, rule-based scoring). $^*$: exact McNemar $p{<}0.05$ vs.\ vanilla; pooled OOD $p{=}0.001$ ($\sigma_0{=}0.2$), $p{<}10^{-4}$ ($\sigma_0{=}0.5$). NC-GRPO is the only intervention in our study improving in-domain, hallucination, and OOD simultaneously; the OOD and in-domain components reproduce on a second seed, hallucination is single-seed (\cref{sec:limitations}).}
  \label{tab:main}
\end{table*}

\Cref{tab:main} (visualized in the supplementary material): both noise scales beat vanilla GRPO on five of six benchmarks with pooled OOD significance. $\sigma_0{=}0.5$ is nominally strongest ($+1.5$pp OOD average, single seed) and individually significant on MathVerse ($p{=}0.0006$) and WeMath ($p{=}0.005$); the two scales are statistically indistinguishable from each other ($p{=}0.13$), so we read the two scales as two operating points at which the effect appears rather than as a ranking; apart from the three configurations for which we trained a second seed (below), each configuration is a single training run, so those tests compare specific checkpoints, not seed distributions (\cref{sec:limitations}). \Cref{fig:dynamics} shows training dynamics: NC-GRPO tracks vanilla's training-reward curve while improving held-out accuracy --- the mechanism is not faster in-domain optimization but convergence to a different, more robust solution, consistent with the agreement account of \cref{sec:agreement}. For external context, comparable RLVR-tuned VLMs train on far larger budgets (OpenVLThinker-7B: 35K SFT + 15K RL samples \cite{deng2025openvlthinker,liu2025noisyrollout}) versus 2.1K samples and no SFT here; absolute accuracies are not cross-protocol comparable given our stricter rule-based scorer (\cref{sec:limitations}).

\paragraph{Seed variance.} We trained a second seed for the three load-bearing configurations. Over two seeds, OOD average is $51.9 \pm 0.1$ (vanilla), $52.8 \pm 0.2$ (\ncgrpo{}, $\sigma_0{=}0.2$), and $50.3 \pm 0.2$ (antithetic); in-domain Geo3K is $52.2 \pm 0.8$, $53.3 \pm 0.6$, and $52.2 \pm 0.8$ (mean $\pm$ half-range, $N{=}2$). The ordering is preserved on both seeds ($+1.0$, $+0.7$ for \ncgrpo{}; $-1.9$, $-1.4$ for antithetic). Geo3K varies far more across seeds than the OOD average (half-range $0.8$ vs.\ $0.1$) and the vanilla/\ncgrpo{} Geo3K ranges overlap; the shift is common to both configurations, so we read the in-domain effect as a paired per-seed delta, not a separation of absolute values.
Pooling per-question records across seeds ($n{=}21{,}342$), \ncgrpo{} improves with $p{=}2.7\times10^{-4}$ and antithetic degrades with $p{=}9.0\times10^{-5}$; the antithetic degradation is significant within each seed ($p{=}0.002$, $p{=}0.013$), the \ncgrpo{} gain significant on seed 1 ($p{=}0.0007$) and trending on seed 2 ($p{=}0.090$). With $N{=}2$ these are variance \emph{indicators}, not distributions.

\begin{table*}[t]
  \centering
  \small
  \begin{tabular}{lcccccc}
    \toprule
    & POPE & MMStar & AI2D & RealWorldQA & ChartQA & Pooled vs.\ vanilla \\
    \midrule
    Vanilla GRPO             & 86.8 & 62.9 & \textbf{81.1} & 65.9 & 80.3 & --- \\
    NoisyRollout (image)     & 86.5 & \textbf{63.9} & 81.0 & 65.6 & \textbf{80.6} & $p{=}0.67$ \\
    NC-GRPO $\sigma_0{=}0.2$ & \textbf{87.0}$^\ddagger$ & 62.6 & 80.8 & \textbf{66.3} & 79.8 & $p{=}0.89$ \\
    NC-GRPO $\sigma_0{=}0.5$ & 85.9 & 63.3 & 79.7 & \textbf{66.3} & 80.4 & $p{=}0.0007$ ($-$) \\
    \bottomrule
  \end{tabular}
  \caption{\textbf{General-capability battery} (accuracy \%; identical rule-based scoring). $^\ddagger$ NC-GRPO $\sigma_0{=}0.2$ beats image noise on POPE ($p{=}0.007$). At $\sigma_0{=}0.2$ NC-GRPO is indistinguishable from vanilla pooled over the battery ($p{=}0.89$) --- reasoning gains at no measurable general-capability cost --- while $\sigma_0{=}0.5$ trades a significant pooled regression ($p{=}0.0007$) for larger reasoning gains: noise scale is a specialization dial.}
  \label{tab:general}
\end{table*}

\begin{table*}[t]
  \centering
  \small
  \begin{tabular}{lccccccc}
    \toprule
    & Geo3K (ID) & Hallusion & MathVista & MathVerse & MathVision & WeMath & OOD avg \\
    \midrule
    NoisyRollout (image)     & 52.1 & 70.7 & 71.5 & 46.5$^*$ & 27.8 & 64.3$^*$ & 56.2$^*$ \\
    $\Delta$ vs.\ vanilla    & $-0.8$ & $+0.5$ & $+0.8$ & $+3.5$ & $+0.7$ & $+15.2$ & $+4.2$ \\
    \bottomrule
  \end{tabular}
  \caption{\textbf{Image-space noise on the reasoning suite} (accuracy \%; same protocol and scorer as \cref{tab:main}; vanilla row there). $^*$: exact McNemar $p{<}0.05$ vs.\ vanilla (MathVerse $p{<}10^{-7}$, WeMath $p{<}10^{-32}$; pooled OOD $p{<}10^{-4}$). NoisyRollout is pixel-space augmentation rather than a latent intervention, so we compare it here: its OOD margin is carried almost entirely by WeMath and MathVerse, and it does not improve in-domain.}
  \label{tab:imagerow}
\end{table*}

\subsection{Mechanism analysis}
\label{sec:mechanism}

\begin{table}[t]
  \centering
  \small
  \setlength{\tabcolsep}{4pt}
  \begin{tabular}{lccc}
    \toprule
    Variant & Geo3K (ID) & WeMath & OOD avg \\
    \midrule
    Vanilla GRPO            & 52.9 & 49.1 & 52.0 \\
    NC-GRPO (indep.\ noise) & \textbf{53.9} & \textbf{50.5} & \textbf{53.0} \\
    \; w/ antithetic pairs  & 52.9 & 40.5 & 50.1$^\dagger$ \\
    \; w/ SPSA steering     & 50.1 & 37.9 & 50.1 \\
    \bottomrule
  \end{tabular}
  \caption{\textbf{Variance reduction hurts.} Antithetic pairing (exact $(+\epsilon, -\epsilon)$ pairs, identical budget, $\sigma_0{=}0.2$) is significantly worse than both independent noise ($p{<}10^{-4}$) and vanilla ($^\dagger$ $p{=}0.006$, pooled OOD McNemar). SPSA steering (\cref{eq:spsa}) on top does not recover the loss ($p{=}0.38$ vs.\ antithetic) despite converging to a stable direction ($\norm{\mu}$ grows $30\times$ over training). The antithetic and SPSA OOD averages coincide at this precision.}
  \label{tab:mechanism}
\end{table}

\paragraph{Independent randomness is the active ingredient.} Antithetic pairing performs exactly the variance cancellation it is designed for --- its training-reward curve is indistinguishable from independent noise and its final in-domain validation matches (52.5\% vs.\ 53.0\%) --- yet OOD accuracy drops significantly ($p{=}0.006$ vs.\ vanilla; \cref{tab:mechanism}). The noise-induced spread in group rewards is therefore not estimator variance but the contrastive signal itself; cancelling it in the group baseline removes precisely the gradient component latent noise adds. This is the causal test of \cref{sec:agreement}: had latent noise helped through another channel, antithetic pairing --- which preserves the marginal noise distribution exactly --- would have preserved the benefit.

We acknowledge an alternative reading: because $(+\epsilon, -\epsilon)$ pairs share an advantage group, their reward contributions partially cancel in the baseline, so the degradation could reflect reduced within-group reward spread rather than reduced exploration diversity. Two observations argue against a pure signal-strength account --- advantages are re-normalized by the group reward standard deviation, compensating uniform shrinkage, and rewards are binary functionals of long stochastic decodes, so first-order cancellation is weak --- and the accounts can be disentangled by placing pairs in separate advantage groups, which we leave to future work. Both readings agree on the operative conclusion: the independent component of the noise, not its budget, carries the benefit.

\paragraph{Direction does not matter; diversity does.} The SPSA probe accumulates the antithetic reward gap into a learned latent direction --- and a consistent direction does exist, as $\norm{\mu}$ grows steadily rather than random-walking --- but steering along it yields no OOD improvement over the antithetic base and degrades in-domain accuracy (\cref{tab:mechanism}). Together the probes rule out both ``noise finds a good direction'' and ``noise magnitude acts as a regularization budget'': what matters is that each rollout sees an independent perturbation.

\paragraph{Noise-scale sweep.} $\sigma_0 \in \{0.2, 0.3, 0.4, 0.5\}$ gives OOD averages $\{53.0, 52.4, 50.7, 53.5\}$: both endpoints are individually significant, but every point except $\sigma_0{=}0.2$ is a single run, so these tests establish that the endpoint checkpoints differ from vanilla --- not that the effect is seed-robust across the sweep. The interior dip at $\sigma_0{=}0.4$ is driven entirely by WeMath (40.2), the benchmark most sensitive to rollout-distribution change in our study (image noise $+15.2$; antithetic $-8.6$); interior structure would need additional seeds.

\begin{table*}[t]
  \centering
  \small
  \setlength{\tabcolsep}{3.5pt}
  \begin{tabular}{llccccccc}
    \toprule
    Model & Method & Geo3K (ID) & HalluBench & MathVista & MathVerse & MathVision & WeMath & OOD avg \\
    \midrule
    \multirow{2}{*}{Qwen2.5-VL-3B}
      & Vanilla GRPO    & 42.8 & 59.6 & \textbf{61.6} & \textbf{36.3} & 23.0 & 44.5 & 45.0 \\
      & \ncgrpo{} (ours) & \textbf{42.9} & \textbf{59.7} & 60.0 & 34.2 & \textbf{24.3} & \textbf{47.0} & 45.0 \\
    \midrule
    \multirow{2}{*}{Qwen2.5-VL-7B}
      & Vanilla GRPO    & 52.9 & 70.2 & \textbf{70.7} & 43.0 & 27.1 & 49.1 & 52.0 \\
      & \ncgrpo{} (ours) & \textbf{53.9} & \textbf{71.3} & 70.1 & \textbf{44.9} & \textbf{28.4} & \textbf{50.5} & \textbf{53.0}$^{*}$ \\
    \midrule
    \multirow{2}{*}{Qwen3-VL-8B-Thinking}
      & Vanilla GRPO    & 63.1 & 73.9 & \textbf{74.4} & 50.0 & 30.4 & 72.0 & 60.1 \\
      & \ncgrpo{} (ours) & \textbf{63.2} & \textbf{74.3} & 73.9 & \textbf{51.3} & \textbf{33.1} & \textbf{74.4} & \textbf{61.4}$^{*}$ \\
    \bottomrule
  \end{tabular}
  \caption{\textbf{\ncgrpo{} across model scales and generations within the Qwen line} (Geometry3K training, identical protocol per model; greedy decoding). $^{*}$~pooled per-question McNemar over $n{=}10{,}038$ OOD questions significant in favor of \ncgrpo{} ($p{=}0.001$ for Qwen2.5-VL-7B; $p{<}10^{-4}$ for Qwen3-VL-8B-Thinking).
  Qwen2.5-VL-3B shows no OOD benefit and is analyzed in \cref{tab:scale}.
  }
  \label{tab:generality}
\end{table*}

\subsection{Image-space noise and the specialization dial}
\label{sec:imagecomparison}

\paragraph{Reasoning benchmarks.} NoisyRollout's image noise (\cref{tab:imagerow}) attains a larger OOD average on the math suite (56.2 vs.\ 53.5), but the margin is carried almost entirely by WeMath ($+15.2$) and MathVerse; it trails vanilla in-domain ($-0.8$) and is beaten by latent noise on HallusionBench, MathVision, and Geo3K, with all three methods nearly indistinguishable in in-domain training dynamics (three-method plot in the supplementary material). The in-domain deficit is within Geo3K's seed-to-seed spread (\cref{sec:results}) and single-seed, so we do not read it as a reliable regression; the published NoisyRollout reports an in-domain \emph{gain} under an LLM-judge protocol \cite{liu2025noisyrollout}, and the flip is consistent with our stricter rule-based scorer and matched 60-step budget, applied identically to all methods. The axis on which image noise clearly gives ground is hallucination --- consistent with the mechanism picture: image noise is a visual augmentation whose benefit concentrates on figure-perception robustness, while latent noise regularizes the joint representation.

\paragraph{Hallucination and general capability.} \Cref{tab:general} probes axes away from math figures. Latent noise at $\sigma_0{=}0.2$ is significantly more robust than image noise on POPE object hallucination ($p{=}0.007$), consistent with its HallusionBench edge --- hallucination is a representation-calibration failure, which latent perturbation targets and pixel perturbation does not. Pooled over the battery, $\sigma_0{=}0.2$ is indistinguishable from vanilla ($p{=}0.89$): reasoning gains at no measurable general-capability cost. The same test reveals the cost of scale: $\sigma_0{=}0.5$, the strongest math-OOD point, shows a significant pooled regression ($p{=}0.0007$; uncorrected, see \cref{sec:limitations}). Noise scale is thus a \emph{specialization dial}: turn it up for reasoning, keep it moderate for a generalist model. Image noise sits between ($p{=}0.67$ pooled) but pays its cost on hallucination instead; the two perturbation families are complementary, and only the latent variant is applicable beyond vision.

\begin{table}[t!b]
  \centering\small
  \setlength{\tabcolsep}{4pt}
  \begin{tabular}{lccc}
    \toprule
    Benchmark & Vanilla & \ncgrpo{} & $p$ \\
    \midrule
    Geo3K (ID)  & 42.8 & \textbf{42.9} & 1.00 \\
    \midrule
    HalluBench  & 59.6 & \textbf{59.7} & 0.60 \\
    MathVista   & \textbf{61.6} & 60.0 & 0.32 \\
    MathVerse   & \textbf{36.3} & 34.2 & 0.014$^{*}$ \\
    MathVision  & 23.0 & \textbf{24.3} & 0.10 \\
    WeMath      & 44.5 & \textbf{47.0} & 0.023$^{*}$ \\
    \midrule
    OOD average & 45.0 & 45.0 & 0.81 \\
    \bottomrule
  \end{tabular}
  \caption{\textbf{Scale ablation: Qwen2.5-VL-3B} (\ncgrpo{} at
  $\sigma_0{=}0.2$ transferred from 7B per \cref{prop:calibration}, which
  guarantees scale \emph{calibration} across widths, not optimality at 3B).
  The 7B OOD benefit is absent: pooled McNemar over $n{=}10{,}038$ paired
  questions is indistinguishable ($p{=}0.81$), in-domain unchanged, and two
  benchmarks move significantly in opposite directions and cancel
  (WeMath $+2.5$, MathVerse $-2.1$) --- \emph{harmless but not beneficial}
  below a capacity threshold. $^{*}$~uncorrected.}
  \label{tab:scale}
\end{table}

\subsection{Scales and model generations}
\label{sec:generality}

A natural concern: a model that already produces long-form reasoning traces natively might gain nothing from latent-noise diversification, its decode-time thinking supplying the trajectory diversity. \Cref{tab:generality} answers this: on Qwen3-VL-8B-Thinking, under the identical protocol (same $\sigma_0{=}0.2$, on a different trainer and inference stack), NC-GRPO improves pooled OOD by $+1.3$pp with $p{<}10^{-4}$ --- stronger than on Qwen2.5-VL, with three individually significant benchmarks and the familiar signature of in-domain tied, gains concentrated OOD. Representational perturbation and decode-time deliberation therefore diversify along different axes.

\paragraph{The effect has a capacity floor.} Qwen2.5-VL-3B (\cref{tab:scale}) returns a null: pooled OOD 45.0 vs.\ 45.0 ($p{=}0.81$), in-domain unchanged ($p{=}1.00$). The null is a cancellation, not a flat response --- WeMath $+2.5$ ($p{=}0.023$), MathVerse $-2.1$ ($p{=}0.014$), the latter reversing sign relative to 7B --- so we do not read 3B as an attenuated 7B effect. Both runs are single-seed, and $\sigma_0{=}0.2$ was transferred rather than re-tuned (\cref{prop:calibration} calibrates scale across widths but does not make it optimal), so the ablation supports only the narrow reading that latent-noise diversification is harmless but not beneficial at this capacity; it is why our recorded predictions for the pending families are stated only above the capacity floor.

All results in this paper are on Qwen-family models, and we make no cross-family claim. We are explicit about why: a single transferred-scale run on a new family cannot distinguish ``does not transfer'' from ``wrong scale'' (\cref{sec:limitations}), so we are evaluating other families with per-family noise-scale sweeps and report nothing until they complete. Point predictions and falsification criteria for these families were recorded before any results were seen, and completed sweeps will be judged against them whichever way they fall.

\section{Limitations}
\label{sec:limitations}

\textbf{Seeds and statistics.} Only the three load-bearing configurations have two seeds; everything else is single-run, so reported spreads ($N{=}2$) are indicators, not distributions, and our paired tests quantify model differences, not seed variance. P-values are uncorrected (${\sim}15$ tests): headline results survive Bonferroni, values in $0.006$--$0.05$ are suggestive, and the antithetic result is confirmatory only pooled across seeds. The in-domain claim rests on the paired per-seed delta (Geo3K seed spread exceeds the effect); the hallucination and general-capability batteries are single-seed.
\textbf{Mechanism scope.} The perturbation displaces only the departure state and first-token distribution (\cref{sec:branching}): we claim no sustained-perturbation mechanism, the propagating (K/V-reaching) variant is untested, and pixel noise perturbs the attended representation more persistently than we do. We also did not run a matched first-token temperature control --- NoisyRollout's temperature study \cite{liu2025noisyrollout} argues against a pure-temperature account, but the control remains open, as does the baseline-cancellation reading of the antithetic result (\cref{sec:mechanism}).
\textbf{Protocol.} Rule-based scoring deflates free-form benchmarks (our NoisyRollout reproduction: 46.5 MathVerse vs.\ 53.2 published under an LLM judge; scorer-insensitive HallusionBench agrees, 70.2 vs.\ 69.8); the deflation is identical across methods, so within-paper comparisons stand. WeMath is outlier-sensitive; read fine differences there cautiously.
\textbf{Scale and family.} The 3B null (\cref{tab:scale}) is confounded with the transferred $\sigma_0$ (\cref{prop:calibration} fixes size, not effect); all results are Qwen-family, with cross-family transfer unestablished and under evaluation via per-family noise-scale sweeps (\cref{sec:generality}).

{
    \small
    \bibliographystyle{ieeenat_fullname}
    \bibliography{main}

\begin{thebibliography}{37}
\providecommand{\natexlab}[1]{#1}
\providecommand{\url}[1]{\texttt{#1}}
\expandafter\ifx\csname urlstyle\endcsname\relax
  \providecommand{\doi}[1]{doi: #1}\else
  \providecommand{\doi}{doi: \begingroup \urlstyle{rm}\Url}\fi

\bibitem[Bai et~al.(2025)Bai, Chen, Liu, Wang, Ge, Song, Dang, Wang, Wang,
  Tang, et~al.]{bai2025qwen25vl}
Shuai Bai, Keqin Chen, Xuejing Liu, Jialin Wang, Wenbin Ge, Sibo Song, Kai
  Dang, Peng Wang, Shijie Wang, Jun Tang, et~al.
\newblock Qwen2.5-vl technical report.
\newblock \emph{arXiv preprint arXiv:2502.13923}, 2025.

\bibitem[Chen et~al.(2024)Chen, Li, Dong, Zhang, Zang, Chen, Duan, Wang, Qiao,
  Lin, and Zhao]{chen2024mmstar}
Lin Chen, Jinsong Li, Xiaoyi Dong, Pan Zhang, Yuhang Zang, Zehui Chen, Haodong
  Duan, Jiaqi Wang, Yu Qiao, Dahua Lin, and Feng Zhao.
\newblock Are we on the right way for evaluating large vision-language models?
\newblock In \emph{Advances in Neural Information Processing Systems}, 2024.

\bibitem[Christiano et~al.(2017)Christiano, Leike, Brown, Martic, Legg, and
  Amodei]{christiano2017deep}
Paul~F Christiano, Jan Leike, Tom~B Brown, Miljan Martic, Shane Legg, and Dario
  Amodei.
\newblock Deep reinforcement learning from human preferences.
\newblock In \emph{Advances in Neural Information Processing Systems}, 2017.

\bibitem[de~Moura and Ullrich(2021)]{demoura2021lean4}
Leonardo de Moura and Sebastian Ullrich.
\newblock The lean 4 theorem prover and programming language.
\newblock In \emph{International Conference on Automated Deduction}, 2021.

\bibitem[Deng et~al.(2025)Deng, Bansal, Yin, Peng, Wang, and
  Chang]{deng2025openvlthinker}
Yihe Deng, Hritik Bansal, Fan Yin, Nanyun Peng, Wei Wang, and Kai-Wei Chang.
\newblock Openvlthinker: An early exploration to complex vision-language
  reasoning via iterative self-improvement.
\newblock \emph{arXiv preprint arXiv:2503.17352}, 2025.

\bibitem[Foret et~al.(2021)Foret, Kleiner, Mobahi, and Neyshabur]{foret2021sam}
Pierre Foret, Ariel Kleiner, Hossein Mobahi, and Behnam Neyshabur.
\newblock Sharpness-aware minimization for efficiently improving
  generalization.
\newblock In \emph{International Conference on Learning Representations}, 2021.

\bibitem[Gal and Ghahramani(2016)]{gal2016dropout}
Yarin Gal and Zoubin Ghahramani.
\newblock Dropout as a bayesian approximation: Representing model uncertainty
  in deep learning.
\newblock In \emph{International Conference on Machine Learning}, 2016.

\bibitem[Guan et~al.(2024)Guan, Liu, Wu, Xian, Li, Liu, Wang, Chen, Huang,
  Yacoob, Manocha, and Zhou]{guan2024hallusionbench}
Tianrui Guan, Fuxiao Liu, Xiyang Wu, Ruiqi Xian, Zongxia Li, Xiaoyu Liu, Xijun
  Wang, Lichang Chen, Furong Huang, Yaser Yacoob, Dinesh Manocha, and Tianyi
  Zhou.
\newblock Hallusionbench: An advanced diagnostic suite for entangled language
  hallucination and visual illusion in large vision-language models.
\newblock In \emph{IEEE/CVF Conference on Computer Vision and Pattern
  Recognition}, 2024.

\bibitem[Guo et~al.(2025)Guo, Yang, Zhang, Song, Zhang, Xu, Zhu, Ma, Wang, Bi,
  et~al.]{deepseekr1}
Daya Guo, Dejian Yang, Haowei Zhang, Junxiao Song, Ruoyu Zhang, Runxin Xu,
  Qihao Zhu, Shirong Ma, Peiyi Wang, Xiao Bi, et~al.
\newblock Deepseek-r1: Incentivizing reasoning capability in llms via
  reinforcement learning.
\newblock \emph{arXiv preprint arXiv:2501.12948}, 2025.

\bibitem[Hao et~al.(2024)Hao, Sukhbaatar, Su, Li, Hu, Weston, and
  Tian]{hao2024coconut}
Shibo Hao, Sainbayar Sukhbaatar, DiJia Su, Xian Li, Zhiting Hu, Jason Weston,
  and Yuandong Tian.
\newblock Training large language models to reason in a continuous latent
  space.
\newblock \emph{arXiv preprint arXiv:2412.06769}, 2024.

\bibitem[Huang et~al.(2025)Huang, Jia, Zhai, Cao, Ye, Zhao, Hu, and
  Lin]{huang2025visionr1}
Wenxuan Huang, Bohan Jia, Zijie Zhai, Shaosheng Cao, Zheyu Ye, Fei Zhao, Yao
  Hu, and Shaohui Lin.
\newblock Vision-r1: Incentivizing reasoning capability in multimodal large
  language models.
\newblock \emph{arXiv preprint arXiv:2503.06749}, 2025.

\bibitem[Jain et~al.(2024)Jain, Chiang, Wen, Kirchenbauer, Chu, Somepalli,
  Bartoldson, Kailkhura, Schwarzschild, Saha, et~al.]{jain2024neftune}
Neel Jain, Ping-yeh Chiang, Yuxin Wen, John Kirchenbauer, Hong-Min Chu,
  Gowthami Somepalli, Brian~R Bartoldson, Bhavya Kailkhura, Avi Schwarzschild,
  Aniruddha Saha, et~al.
\newblock Neftune: Noisy embeddings improve instruction finetuning.
\newblock In \emph{International Conference on Learning Representations}, 2024.

\bibitem[Jerge and Evans(2026)]{jerge2026noisycoconut}
Michael~M. Jerge and David Evans.
\newblock Noisycoconut: Counterfactual consensus via latent space reasoning.
\newblock \emph{Transactions on Machine Learning Research}, 2026.

\bibitem[Kembhavi et~al.(2016)Kembhavi, Salvato, Kolve, Seo, Hajishirzi, and
  Farhadi]{kembhavi2016ai2d}
Aniruddha Kembhavi, Mike Salvato, Eric Kolve, Minjoon Seo, Hannaneh Hajishirzi,
  and Ali Farhadi.
\newblock A diagram is worth a dozen images.
\newblock In \emph{European Conference on Computer Vision}, 2016.

\bibitem[Kornblith et~al.(2019)Kornblith, Norouzi, Lee, and
  Hinton]{kornblith2019cka}
Simon Kornblith, Mohammad Norouzi, Honglak Lee, and Geoffrey Hinton.
\newblock Similarity of neural network representations revisited.
\newblock In \emph{International Conference on Machine Learning}, 2019.

\bibitem[Lakshminarayanan et~al.(2017)Lakshminarayanan, Pritzel, and
  Blundell]{lakshminarayanan2017ensembles}
Balaji Lakshminarayanan, Alexander Pritzel, and Charles Blundell.
\newblock Simple and scalable predictive uncertainty estimation using deep
  ensembles.
\newblock In \emph{Advances in Neural Information Processing Systems}, 2017.

\bibitem[Li et~al.(2023)Li, Du, Zhou, Wang, Zhao, and Wen]{li2023pope}
Yifan Li, Yifan Du, Kun Zhou, Jinpeng Wang, Wayne~Xin Zhao, and Ji-Rong Wen.
\newblock Evaluating object hallucination in large vision-language models.
\newblock In \emph{Conference on Empirical Methods in Natural Language
  Processing}, 2023.

\bibitem[Liang et~al.(2021)Liang, Wu, Li, Wang, Meng, Qin, Chen, Zhang, and
  Liu]{liang2021rdrop}
Xiaobo Liang, Lijun Wu, Juntao Li, Yue Wang, Qi Meng, Tao Qin, Wei Chen, Min
  Zhang, and Tie-Yan Liu.
\newblock R-drop: Regularized dropout for neural networks.
\newblock In \emph{Advances in Neural Information Processing Systems}, 2021.

\bibitem[Liu et~al.(2025{\natexlab{a}})Liu, Ni, Wu, Du, Dou, Wang, Pang, and
  Shieh]{liu2025noisyrollout}
Xiangyan Liu, Jinjie Ni, Zijian Wu, Chao Du, Longxu Dou, Haonan Wang, Tianyu
  Pang, and Michael~Qizhe Shieh.
\newblock Noisyrollout: Reinforcing visual reasoning with data augmentation.
\newblock In \emph{Advances in Neural Information Processing Systems},
  2025{\natexlab{a}}.

\bibitem[Liu et~al.(2025{\natexlab{b}})Liu, Chen, Li, Qi, Pang, Du, Lee, and
  Lin]{liu2025drgrpo}
Zichen Liu, Changyu Chen, Wenjun Li, Penghui Qi, Tianyu Pang, Chao Du, Wee~Sun
  Lee, and Min Lin.
\newblock Understanding r1-zero-like training: A critical perspective.
\newblock \emph{arXiv preprint arXiv:2503.20783}, 2025{\natexlab{b}}.

\bibitem[Lu et~al.(2021)Lu, Gong, Jiang, Qiu, Huang, Liang, and
  Zhu]{lu2021intergps}
Pan Lu, Ran Gong, Shibiao Jiang, Liang Qiu, Siyuan Huang, Xiaodan Liang, and
  Song-Chun Zhu.
\newblock Inter-gps: Interpretable geometry problem solving with formal
  language and symbolic reasoning.
\newblock In \emph{Annual Meeting of the Association for Computational
  Linguistics}, 2021.

\bibitem[Lu et~al.(2024)Lu, Bansal, Xia, Liu, Li, Hajishirzi, Cheng, Chang,
  Galley, and Gao]{lu2024mathvista}
Pan Lu, Hritik Bansal, Tony Xia, Jiacheng Liu, Chunyuan Li, Hannaneh
  Hajishirzi, Hao Cheng, Kai-Wei Chang, Michel Galley, and Jianfeng Gao.
\newblock Mathvista: Evaluating mathematical reasoning of foundation models in
  visual contexts.
\newblock In \emph{International Conference on Learning Representations}, 2024.

\bibitem[Masry et~al.(2022)Masry, Long, Tan, Joty, and Hoque]{masry2022chartqa}
Ahmed Masry, Do~Xuan Long, Jia~Qing Tan, Shafiq Joty, and Enamul Hoque.
\newblock Chartqa: A benchmark for question answering about charts with visual
  and logical reasoning.
\newblock In \emph{Findings of the Association for Computational Linguistics},
  2022.

\bibitem[Meng et~al.(2025)Meng, Du, Liu, Zhou, Lu, Fu, Shi, Wang, He, Zhang,
  et~al.]{meng2025mmeureka}
Fanqing Meng, Lingxiao Du, Zongkai Liu, Zhixiang Zhou, Quanfeng Lu, Daocheng
  Fu, Botian Shi, Wenhai Wang, Junjun He, Kaipeng Zhang, et~al.
\newblock Mm-eureka: Exploring visual aha moment with rule-based large-scale
  reinforcement learning.
\newblock \emph{arXiv preprint arXiv:2503.07365}, 2025.

\bibitem[Ouyang et~al.(2022)Ouyang, Wu, Jiang, Almeida, Wainwright, Mishkin,
  Zhang, Agarwal, Slama, Ray, et~al.]{ouyang2022instructgpt}
Long Ouyang, Jeffrey Wu, Xu Jiang, Diogo Almeida, Carroll Wainwright, Pamela
  Mishkin, Chong Zhang, Sandhini Agarwal, Katarina Slama, Alex Ray, et~al.
\newblock Training language models to follow instructions with human feedback.
\newblock In \emph{Advances in Neural Information Processing Systems}, 2022.

\bibitem[Qiao et~al.(2025)Qiao, Tan, Dong, Wu, Sun, Song, GongQue, Lei, Wei,
  Zhang, et~al.]{qiao2024wemath}
Runqi Qiao, Qiuna Tan, Guanting Dong, Minhui Wu, Chong Sun, Xiaoshuai Song,
  Zhuoma GongQue, Shanglin Lei, Zhe Wei, Miaoxuan Zhang, et~al.
\newblock We-math: Does your large multimodal model achieve human-like
  mathematical reasoning?
\newblock In \emph{Annual Meeting of the Association for Computational
  Linguistics}, 2025.

\bibitem[Salimans et~al.(2017)Salimans, Ho, Chen, Sidor, and
  Sutskever]{salimans2017es}
Tim Salimans, Jonathan Ho, Xi Chen, Szymon Sidor, and Ilya Sutskever.
\newblock Evolution strategies as a scalable alternative to reinforcement
  learning.
\newblock \emph{arXiv preprint arXiv:1703.03864}, 2017.

\bibitem[Schulman et~al.(2017)Schulman, Wolski, Dhariwal, Radford, and
  Klimov]{schulman2017ppo}
John Schulman, Filip Wolski, Prafulla Dhariwal, Alec Radford, and Oleg Klimov.
\newblock Proximal policy optimization algorithms.
\newblock \emph{arXiv preprint arXiv:1707.06347}, 2017.

\bibitem[Shao et~al.(2024)Shao, Wang, Zhu, Xu, Song, Bi, Zhang, Zhang, Li, Wu,
  and Guo]{shao2024deepseekmath}
Zhihong Shao, Peiyi Wang, Qihao Zhu, Runxin Xu, Junxiao Song, Xiao Bi, Haowei
  Zhang, Mingchuan Zhang, YK Li, Yang Wu, and Daya Guo.
\newblock Deepseekmath: Pushing the limits of mathematical reasoning in open
  language models.
\newblock \emph{arXiv preprint arXiv:2402.03300}, 2024.

\bibitem[Spall(1992)]{spall1992spsa}
James~C Spall.
\newblock Multivariate stochastic approximation using a simultaneous
  perturbation gradient approximation.
\newblock \emph{IEEE Transactions on Automatic Control}, 37\penalty0
  (3):\penalty0 332--341, 1992.

\bibitem[{The mathlib Community}(2020)]{mathlib2020}
{The mathlib Community}.
\newblock The lean mathematical library.
\newblock In \emph{Proceedings of the 9th ACM SIGPLAN International Conference
  on Certified Programs and Proofs}, 2020.

\bibitem[von Klinski and Schall(2026)]{vonklinski2026taxonrl}
Maximilian von Klinski and Maximilian Schall.
\newblock Taxonrl: Reinforcement learning with intermediate rewards for
  interpretable fine-grained visual reasoning.
\newblock \emph{arXiv preprint arXiv:2603.04380}, 2026.

\bibitem[Wang et~al.(2024)Wang, Pan, Shi, Lu, Zhan, and Li]{wang2024mathvision}
Ke Wang, Junting Pan, Weikang Shi, Zimu Lu, Mingjie Zhan, and Hongsheng Li.
\newblock Measuring multimodal mathematical reasoning with math-vision dataset.
\newblock In \emph{Advances in Neural Information Processing Systems}, 2024.

\bibitem[{xAI}(2024)]{xai2024realworldqa}
{xAI}.
\newblock Realworldqa: A benchmark for real-world spatial understanding.
\newblock \url{https://huggingface.co/datasets/xai-org/RealworldQA}, 2024.

\bibitem[Yu et~al.(2025{\natexlab{a}})Yu, Xu, and Chen]{yu2025onpolicy}
Chengzhi Yu, Yifan Xu, and Yifan Chen.
\newblock Optimizing lvlms with on-policy data for effective hallucination
  mitigation.
\newblock \emph{arXiv preprint arXiv:2512.00706}, 2025{\natexlab{a}}.

\bibitem[Yu et~al.(2025{\natexlab{b}})Yu, Zhang, Zhu, Yuan, Zuo, Yue, Fan, Liu,
  Liu, Liu, et~al.]{yu2025dapo}
Qiying Yu, Zheng Zhang, Ruofei Zhu, Yufeng Yuan, Xiaochen Zuo, Yu Yue, Tiantian
  Fan, Gaohong Liu, Lingjun Liu, Xin Liu, et~al.
\newblock Dapo: An open-source llm reinforcement learning system at scale.
\newblock In \emph{Advances in Neural Information Processing Systems},
  2025{\natexlab{b}}.

\bibitem[Zhang et~al.(2024)Zhang, Jiang, Zhang, Lin, Guo, Qiu, Zhou, Lu, Chang,
  Gao, and Li]{zhang2024mathverse}
Renrui Zhang, Dongzhi Jiang, Yichi Zhang, Haokun Lin, Ziyu Guo, Pengshuo Qiu,
  Aojun Zhou, Pan Lu, Kai-Wei Chang, Peng Gao, and Hongsheng Li.
\newblock Mathverse: Does your multi-modal llm truly see the diagrams in visual
  math problems?
\newblock In \emph{European Conference on Computer Vision}, 2024.

\end{thebibliography}
}

\clearpage
\appendix
\section{Proofs}
\label{sec:proofs}

The algebraic content of these proofs is machine-verified in Lean~4 \cite{demoura2021lean4} with Mathlib \cite{mathlib2020} (included in the released artifact): the full advantage decomposition of \cref{prop:decomposition} including the zero-mean property of the contrast term; the norm-scaling reduction and mean/variance arithmetic of \cref{prop:calibration}, with the $\chi^2_d$ moments entering as hypotheses; the even/odd decomposition, exact pair-mean cancellation, and oddness of the first-order term in \cref{lem:antithetic}; and the trace identity $\E[\epsilon^\top M \epsilon] = \tr M$ used in \cref{eq:smoothing}, with the second-moment structure as hypotheses. The verification covers the algebraic steps only; Gaussian moment computations, the Taylor remainder bound in \cref{eq:smoothing}, and the first-order expansion for the pixel-noise geometry remain pencil-and-paper.

\begin{proofof}{\cref{prop:calibration}}
With $\tilde{h} - h = \sigma \frac{\norm{h}}{\sqrt{d}} \epsilon$ and $\epsilon \sim \mathcal{N}(0, I_d)$,
\begin{equation*}
\big\| \tilde{h} - h \big\|^2 = \frac{\sigma^2 \norm{h}^2}{d} \norm{\epsilon}^2, \qquad \norm{\epsilon}^2 \sim \chi^2_d
\end{equation*}
with $\E \norm{\epsilon}^2 = d$ and $\Var \norm{\epsilon}^2 = 2d$. Hence $\E \| \tilde{h} - h \|^2 = \sigma^2 \norm{h}^2$ and $\Var\big( \| \tilde{h} - h \|^2 / (\sigma^2 \norm{h}^2) \big) = \Var\big( \norm{\epsilon}^2 / d \big) = 2/d$.
\end{proofof}

\begin{proofof}{\cref{prop:decomposition}}
The group mean over the $2n$ rollouts is $\bar{r} = \frac{1}{2n}\big( n \bar{r}^{c} + n \bar{r}^{\eta} \big) = \frac{1}{2}(\bar{r}^{c} + \bar{r}^{\eta})$. For a clean rollout,
\begin{align*}
\hat{A}_i^{c} = \frac{r_i^{c} - \bar{r}}{s}
  &= \frac{r_i^{c} - \bar{r}^{c}}{s} + \frac{\bar{r}^{c} - \frac{1}{2}(\bar{r}^{c} + \bar{r}^{\eta})}{s} \\
  &= \frac{r_i^{c} - \bar{r}^{c}}{s} + \frac{\bar{r}^{c} - \bar{r}^{\eta}}{2s},
\end{align*}
and symmetrically for a noisy rollout with the sign of the contrast term reversed. The contrast term has zero mean over the group (the two halves contribute $+n$ and $-n$ copies), and its population value $\frac{1}{2s}\big( \E[r \mid \text{clean}] - \E[r \mid \text{noisy}] \big)$ vanishes iff expected reward is invariant to the perturbation.
\end{proofof}

\paragraph{Noise-smoothed reward and curvature (\cref{eq:smoothing}).}
\emph{Let $\varphi(h) = \E[\, r \mid \text{prefill state } h \,]$ (decode randomness integrated out) be three times continuously differentiable with uniformly bounded third derivatives. Under the calibrated perturbation of \cref{eq:calibrated} with scale $\sigma$, the expansion of \cref{eq:smoothing} holds, with $\bar{\lambda}(\nabla^2 \varphi) = \frac{1}{d} \tr \nabla^2 \varphi$ the mean eigenvalue of the Hessian.}
\smallskip

\begin{proofof}{\cref{eq:smoothing}}
Let $a = \sigma \norm{h} / \sqrt{d}$, so $\tilde{h} = h + a\epsilon$. A third-order Taylor expansion with integral remainder gives
\begin{equation*}
\varphi(h + a\epsilon) = \varphi(h) + a\, \nabla \varphi(h)^\top \epsilon + \frac{a^2}{2}\, \epsilon^\top \nabla^2 \varphi(h)\, \epsilon + R(\epsilon),
\end{equation*}
with $|R(\epsilon)| \le \frac{C}{6} a^3 \norm{\epsilon}^3$ for $C = \sup_h \| \nabla^3 \varphi(h) \|$. Taking expectations: $\E[\epsilon] = 0$ kills the linear term; $\E[\epsilon^\top M \epsilon] = \tr M$ gives $\frac{a^2}{2} \tr \nabla^2 \varphi(h) = \frac{\sigma^2 \norm{h}^2}{2} \cdot \frac{\tr \nabla^2 \varphi(h)}{d}$; and $\E \norm{\epsilon}^3 \le \big( \E \norm{\epsilon}^4 \big)^{3/4} = O(d^{3/2})$ bounds the remainder by $O(C \sigma^3 \norm{h}^3)$, which is $O(\sigma^3)$ uniformly in $d$ after the $a = \sigma \norm{h} / \sqrt{d}$ substitution.
\end{proofof}

\begin{proofof}{\cref{lem:antithetic}}
Since $-\epsilon \overset{d}{=} \epsilon$, each member of an antithetic pair is marginally distributed as an independent draw, so per-rollout statistics are unchanged. Write $\varphi_{\mathrm{even}}(\epsilon) = \frac{1}{2}(\varphi(\epsilon) + \varphi(-\epsilon))$ and $\varphi_{\mathrm{odd}}(\epsilon) = \frac{1}{2}(\varphi(\epsilon) - \varphi(-\epsilon))$; then the pair mean is exactly $\varphi_{\mathrm{even}}(\epsilon)$, so the noisy-half mean over $n/2$ pairs is an average of iid copies of $\varphi_{\mathrm{even}}$ and carries no odd-component fluctuation. The first-order Taylor term $a\, \nabla \varphi(h)^\top \epsilon$ is odd in $\epsilon$, hence removed; the quadratic term is even, hence retained, and by \cref{eq:smoothing} its expectation is the $O(\sigma^2)$ curvature term. Because $\E[\varphi_{\mathrm{even}} \varphi_{\mathrm{odd}}]$ integrates an odd function against the symmetric Gaussian, the even and odd parts are uncorrelated, so for small $\sigma$ (where $\Var \varphi_{\mathrm{odd}} = a^2 \norm{\nabla \varphi}^2 + O(\sigma^4)$ dominates $\Var \varphi_{\mathrm{even}} = O(\sigma^4)$), antithetic pairing removes the dominant share of the noisy-half mean's variance.
\end{proofof}

\paragraph{Induced geometry of pixel noise (\cref{sec:geometry}).}
\emph{Pixel noise $\delta \sim \mathcal{N}(0, s^2 I_N)$ induces, to first order, a representation perturbation $\Delta h = J\delta + O(\norm{\delta}^2)$ that is Gaussian with covariance $s^2 J J^\top$. Consequently: (i) its support is $\operatorname{range}(J)$; (ii) its power along a unit direction $u$ is $s^2 \| J^\top u \|^2$; (iii) its total power $\E \norm{\Delta h}^2 = s^2 \norm{J(x)}_F^2$ is input-dependent; and (iv) its second-order smoothing penalty is $\frac{s^2}{2} \tr \nabla^2 (\varphi \circ f)$ --- reward curvature filtered through the encoder geometry.}
\smallskip

\begin{proofof}{the induced geometry of pixel noise}
A first-order expansion gives $\Delta h = f(x + \delta) - f(x) = J\delta + O(\norm{\delta}^2)$; linear images of Gaussians are Gaussian, so to first order $\Delta h \sim \mathcal{N}(0, s^2 J J^\top)$. (i) $J\delta \in \operatorname{range}(J)$ for every $\delta$, and for $u \perp \operatorname{range}(J)$, $\Var(u^\top J \delta) = s^2 \| J^\top u \|^2 = 0$. (ii) Direct computation: $\Var(u^\top \Delta h) = s^2 u^\top J J^\top u = s^2 \| J^\top u \|^2$, which ranges over the squared singular values of $J$ as $u$ ranges over the corresponding singular directions. (iii) $\E \norm{J\delta}^2 = s^2 \tr(J J^\top) = s^2 \norm{J}_F^2$, an input-dependent quantity. (iv) Apply the argument behind \cref{eq:smoothing} to the composition $\varphi \circ f$: the second-order coefficient is $\frac{s^2}{2} \tr \nabla^2 (\varphi \circ f)(x)$, and the chain rule gives $\nabla^2 (\varphi \circ f) = J^\top \nabla^2 \varphi\, J + \sum_k \partial_k \varphi\, \nabla^2 f_k$.
\end{proofof}

\section{Workflow, Training-Step Algorithm, and Additional Plots}
\label{sec:suppthreemethod}

\begin{figure*}[t]
  \centering
  \resizebox{0.8\textwidth}{!}{
\begin{tikzpicture}[
  font=\footnotesize,
  node distance=0.45cm and 0.7cm,
  box/.style={draw, rounded corners=2pt, align=center, inner sep=4pt,
              minimum height=0.85cm},
  clean/.style={box, fill=black!6, draw=black!60},
  noisy/.style={box, fill=wacvblue!12, draw=wacvblue},
  arr/.style={-{Stealth[length=2mm]}, semithick},
  narr/.style={arr, wacvblue, dashed},
  lbl/.style={font=\scriptsize, align=center}
]
\node[box] (input) {input image $I$\\prompt $q$};

\node[clean, above right=0.35cm and 0.7cm of input]
  (cprefill) {clean prefill\\hidden states $h$};
\node[noisy, below right=0.35cm and 0.7cm of input]
  (nprefill) {noisy prefill (last layer)\\$\tilde{h} = h + \sigma_k \tfrac{\norm{h}}{\sqrt{d}}\,\epsilon$};
\node[lbl, wacvblue, below=0.15cm of nprefill]
  (eps) {$\epsilon \sim \mathcal{N}(0, I_d)$ seeded; $\sigma_k \to 0$ (\cref{eq:schedule})};

\node[clean, right=of cprefill] (cdec) {decode $n$ rollouts\\$o_1, \ldots, o_n$};
\node[noisy, right=of nprefill] (ndec) {decode $n$ rollouts\\$o_{n+1}, \ldots, o_{2n}$};
\node[lbl, below=0.15cm of ndec] {no decode-time noise};

\node[box] (adv) at ($(cdec.east)!0.5!(ndec.east) + (2.55,0)$)
  {verifiable rewards $r_i$\\[1pt] joint-group advantages\\$\hat{A}_i = (r_i - \bar{r})/s$};
\node[lbl, below=0.15cm of adv] {clean/noisy contrast\\term (\cref{prop:decomposition})};

\node[box, right=0.7cm of adv] (upd) {GRPO update\\(\cref{eq:grpo})\\noise-free forward};

\draw[arr]  (input.east) -- ++(0.32,0) |- (cprefill.west);
\draw[narr] (input.east) -- ++(0.32,0) |- (nprefill.west);
\draw[arr]  (cprefill) -- (cdec);
\draw[narr] (nprefill) -- (ndec);
\draw[arr]  (cdec.east)  -- ++(0.32,0) |- ([yshift=2mm]adv.west);
\draw[narr] (ndec.east)  -- ++(0.32,0) |- ([yshift=-2mm]adv.west);
\draw[arr]  (adv) -- (upd);
\end{tikzpicture}}
  \caption{The NC-GRPO workflow. Each prompt is prefilled twice: once clean (solid, gray) and once with scale-calibrated Gaussian noise added to the hidden states returned by the last layer (dashed, blue). The perturbation shifts the logits at the final prompt position, so each noisy branch departs from a displaced state and then decodes from a clean cache with no decode-time noise (\cref{sec:branching}); rewards are normalized over the joint $2n$-rollout group, so systematic disagreement between the clean and noisy halves --- the contrast term of \cref{prop:decomposition} --- becomes learning signal. The noise scale $\sigma_k$ anneals to zero over training (\cref{eq:schedule}), and the policy-gradient forward pass sees only unperturbed states.}
  \label{fig:workflow}
\end{figure*}
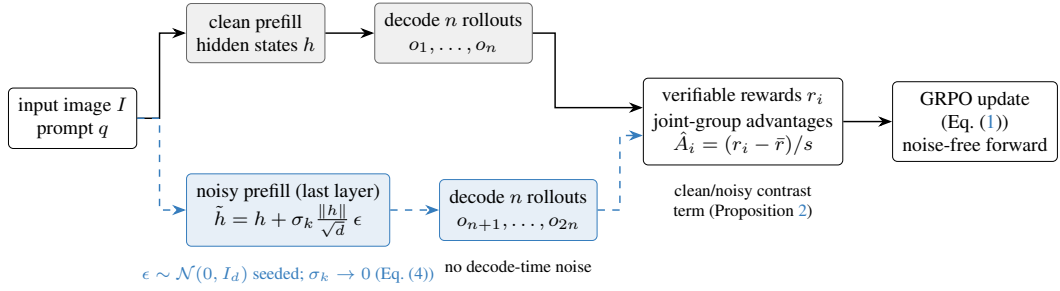

\begin{figure}[t]
  \centering
  \includegraphics[width=0.8\linewidth]{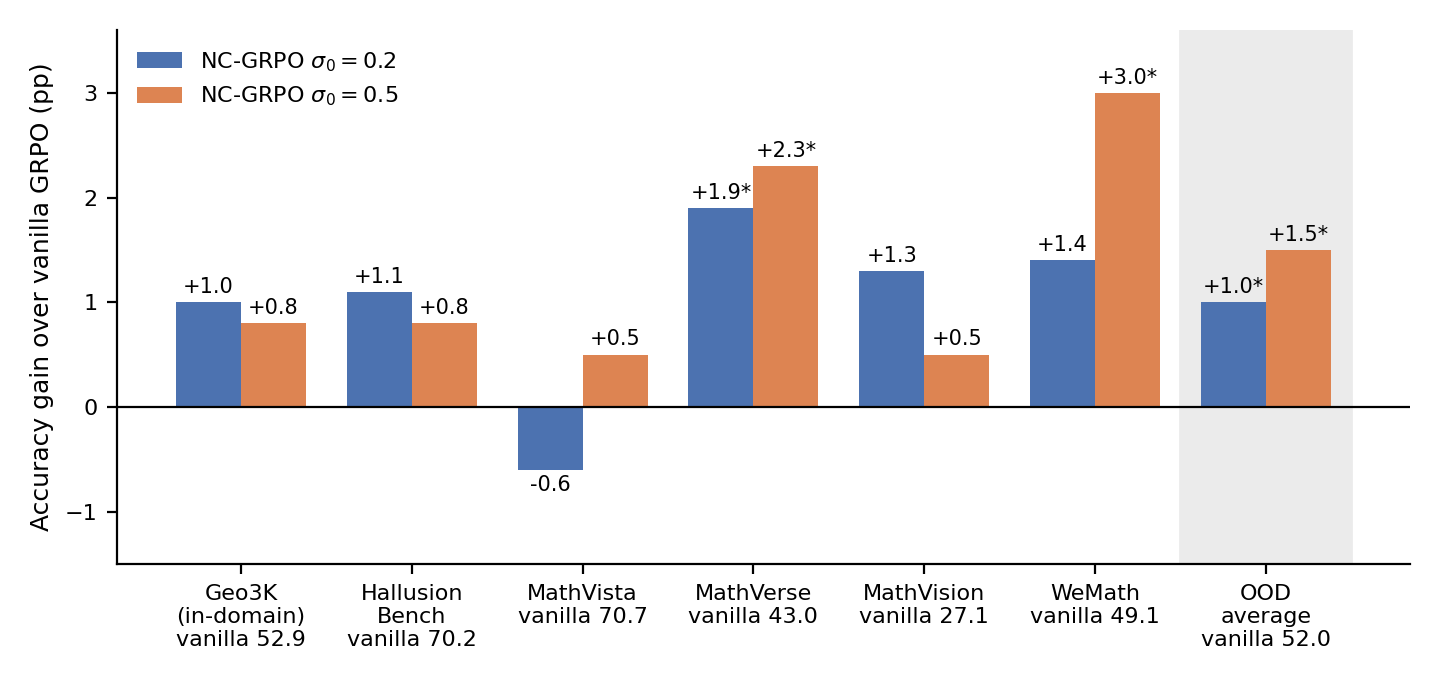}
  \caption{NC-GRPO vs.\ vanilla GRPO at both noise scales ($\sigma_0 \in \{0.2, 0.5\}$): accuracy gains in percentage points, with vanilla's absolute accuracy beneath each benchmark; visualization of the main-results table in the main paper. Shaded column: OOD average over the five held-out benchmarks. $^*$: exact McNemar $p{<}0.05$ on paired per-question outcomes.}
  \label{fig:gains}
\end{figure}

\begin{figure}[t]
  \centering
  \includegraphics[width=0.8\linewidth]{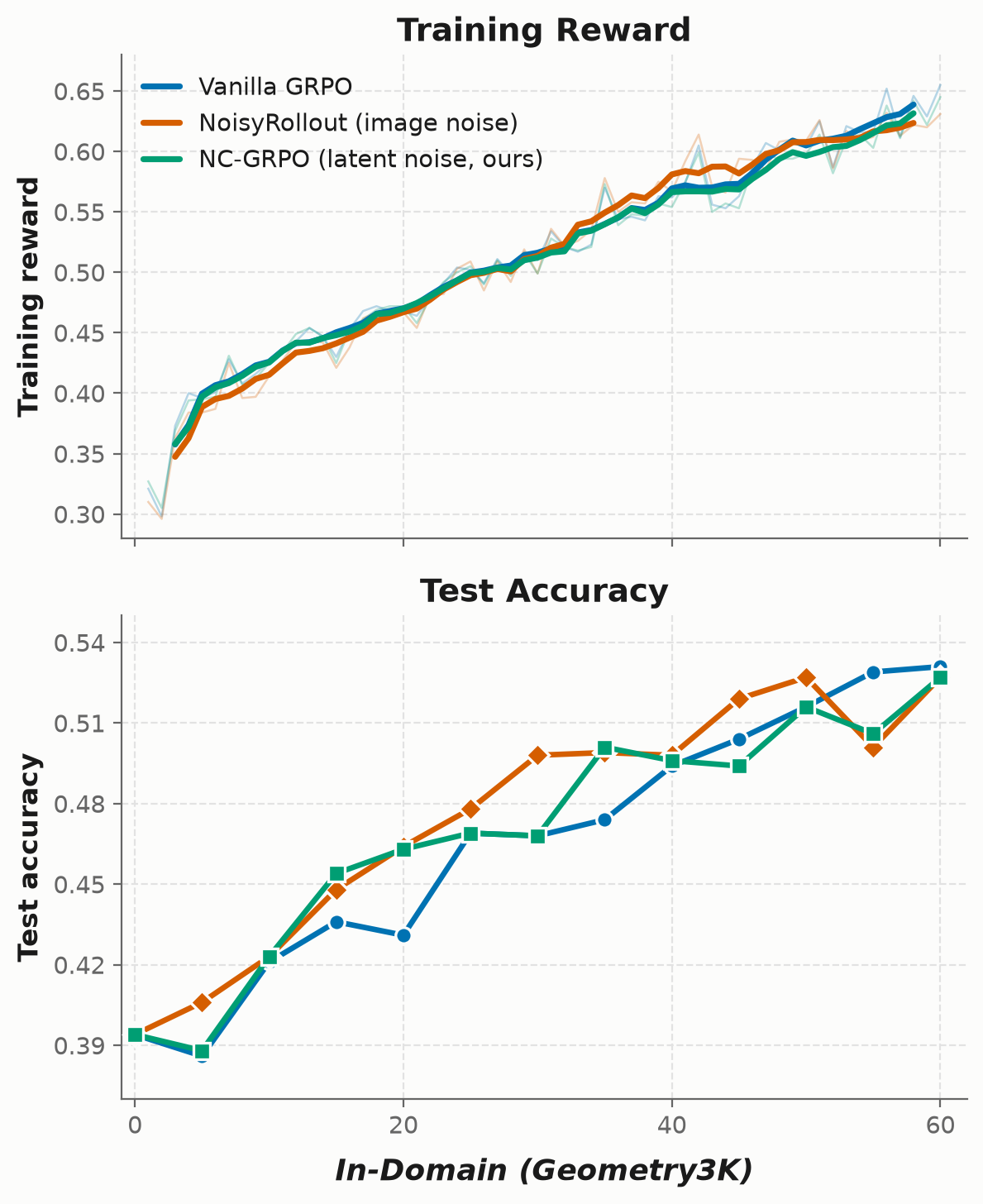}
  \caption{In-domain dynamics of all three methods (Geometry3K): training reward (top) and validation accuracy (bottom) for vanilla GRPO, NoisyRollout (image noise), and NC-GRPO (latent noise). The three training-reward curves are nearly indistinguishable and in-domain validation separates only marginally --- the methods differ OOD (\cref{tab:main,tab:imagerow}), not in-domain. Referenced from the image-comparison subsection of the main paper.}
  \label{fig:threemethod}
\end{figure}

\begin{algorithm}[t]
\caption{NC-GRPO training step (minimal instantiation)}
\label{alg:ncgrpo}
\begin{algorithmic}[1]
\Require policy $\pi_\theta$, old policy $\pi_{\theta_{\mathrm{old}}}$, prompt $q$ with verifiable answer $y^*$, step $k$, group size $2n$
\State $\sigma_k \gets$ schedule of \cref{eq:schedule}
\State $\{o_i\}_{i=1}^{n} \gets$ sample $n$ rollouts from clean prefill
\For{$i = n{+}1, \ldots, 2n$} \Comment{noisy half}
    \State draw $\epsilon_t \sim \mathcal{N}(0, I_d)$ per prompt token (seeded)
    \State perturb the \emph{returned} last-layer prefill states via \cref{eq:calibrated}
    \Statex \quad\ \ \textit{(KV cache is written from clean activations)}
    \State sample first token of $o_i$ from the perturbed logits
    \State decode the remainder of $o_i$ from the clean cache (no decode noise)
\EndFor
\State $r_i \gets \mathbf{1}[\operatorname{extract}(o_i) = y^*]$ for all $i$
\State $\hat{A}_i \gets (r_i - \bar{r}) / (\operatorname{std}(r) + \delta)$ over the joint group
\State update $\theta$ with \cref{eq:grpo} \Comment{noise-free forward}
\end{algorithmic}
\end{algorithm}

\end{document}